\documentclass{bmvc2k}

\usepackage{multirow}
\usepackage{amsmath}
\usepackage{xcolor, colortbl}
\usepackage{sidecap}

\usepackage{pifont}
\newcommand{\bcmark}{\ding{51}}%
\newcommand{\cmark}{\textcolor[rgb]{0,0.8,0}{\ding{51}}}%
\newcommand{\xmark}{\textcolor{red}{\ding{55}}}%

\newcommand{\highlight}[2][yellow]{\mathchoice%
  {\colorbox{#1}{$\displaystyle#2$}}%
  {\colorbox{#1}{$\textstyle#2$}}%
  {\colorbox{#1}{$\scriptstyle#2$}}%
  {\colorbox{#1}{$\scriptscriptstyle#2$}}}%

\graphicspath{{images/}}
  
\sidecaptionvpos{figure}{c}

\definecolor{mygreen}{rgb}{0.88,0.94,0.85}
\definecolor{myred}{rgb}{0.98,0.89,0.84}
\definecolor{myyellow}{rgb}{1,0.94,0.80}
\definecolor{Gray}{gray}{0.85}

\title{Residual Conv-Deconv Grid Network for Semantic Segmentation}
\newcommand{\gridnet}{GridNet\xspace}
\newcommand{\gridnets}{GridNets\xspace}
\newcommand{\cnn}{CNN\xspace}

\addauthor{Damien Fourure}{damien.fourure@univ-st-etienne.fr}{1}
\addauthor{R\'emi Emonet}{remi.emonet@univ-st-etienne.fr}{1}
\addauthor{Elisa Fromont}{elisa.fromont@univ-st-etienne.fr}{1}
\addauthor{Damien Muselet}{damien.muselet@univ-st-etienne.fr}{1}
\addauthor{Alain Tremeau}{alain.tremeau@univ-st-etienne.fr}{1}
\addauthor{Christian Wolf}{christian.wolf@liris.cnrs.fr}{2}

\addinstitution{
Univ Lyon,  UJM Saint-Etienne, \\
CNRS UMR 5516,\\
 Hubert Curien Lab, F-42023\\
 Saint-Etienne,  France
}
\addinstitution{
INSA-Lyon,\\
LIRIS UMR CNRS 5205, \\
F-69621, \\
France
}

\runninghead{Fourure \etal}{Residual Conv-Deconv \gridnet}

\def\etal{\emph{et al}\bmvaOneDot}

\begin{document}

\maketitle

\begin{abstract}
\noindent
This paper presents \gridnet, a new Convolutional Neural Network (\cnn) architecture for semantic image segmentation (full scene labelling).
Classical neural networks are implemented as one stream from the input to the output with subsampling operators applied in the stream in order to reduce the feature maps size and to increase the receptive field for the final prediction.
However, for semantic image segmentation, where the task consists in providing a semantic class to each pixel of an image, feature maps reduction is harmful because it leads to a resolution loss in the output prediction.  
To tackle this problem, our \gridnet follows a grid pattern allowing multiple interconnected streams to work at different resolutions. 
We show that our network generalizes many well known networks such as conv-deconv, residual or U-Net networks. \gridnet is trained from scratch and achieves competitive results on the Cityscapes dataset. 
\end{abstract}


\section{Introduction}
\label{sec:intro}

\noindent
Convolutional Neural Networks (\cnn) have become tremendously popular for a huge number of applications \cite{Pham2016ConvolutionalNN, ShahroudyNYW16, BaiSZCQ17} since the success of AlexNet \cite{KrizhevskySH12} in 2012.
AlexNet, VGG16 \cite{VGG16} and ResNet \cite{residual}, are some of the famous architectures designed for image classification which have shown incredible results. 
While image classification aims at predicting a single class per image (presence or not of an object in an image) we tackle the problem of full scene labelling. Full scene labelling or semantic segmentation from RGB images aims at segmenting an image into semantically meaningful regions, i.e. at providing a class label for each pixel of an image.
Based on the success of classical \cnn, new networks designed especially for semantic segmentation, named fully convolutional networks have been developed. 
The main advantage of these networks is that they produce 2D matrices as output, allowing the network to label an entire image directly. Because they are fully convolutional, they can be fed with images of various sizes. 

In order to construct fully convolutional networks, two strategies have been developed: conv-deconv networks and dilated convolution-based networks (see Section~\ref{sec:related} for more details). 
Conv-deconv networks are composed of two parts: the first one is a classical convolutional network with subsampling operations which decrease the feature maps sizes and the second part is a deconvolutional network with upsampling operations which increase the feature maps sizes back to the original input resolution.
Dilated convolution-based networks~\cite{dilated_convolution} do not use subsampling operations but a "\`a trous" algorithm on dilated convolutions to increase the receptive field of the network.

If increasing the depth of the network has often gone hand in hand with increasing the performance on many data rich applications, it has also been observed that the deeper the network, the more difficult its training is, due to vanishing gradient problems during the back-propagation steps. Residual networks \cite{residual} (ResNet) solve this problem by using identity residual connections to allow the gradient to back-propagate more easily. As a consequence, they are often faster to train than classical neural networks. The residual connections are thus now commonly used in all new architectures. 

Lots of pre-trained (usually on Imagenet \cite{imagenet}) ResNet are available for the community. They can be fine-tuned for a new task. However, the structure of a pre-trained network cannot be changed radically which is a problem when a new architecture, such as ours, comes out.
  

In this paper we present \gridnet, a new architecture especially designed for full scene labelling. \gridnet is composed of multiple paths from the input image to the output prediction, that we call streams, working at different image resolutions.
High resolution streams allow the network to give an accurate prediction in combination with low resolution streams which carry more context thanks to bigger receptive fields. The streams are interconnected with convolutional and deconvolutional units to form the columns of our grid. With these connections, information from low and high resolutions can be shared.

In Section~\ref{sec:related}, we review the network architectures used for full scene labelling from which \gridnet takes inspiration and we show how our approach generalises existing methods.
In Section~\ref{sec:gridnet}, we present the core components of the proposed \gridnet architecture.
Finally, Section~\ref{sec:exp} shows results on the Cityscapes dataset.

\section{Related Work}
\label{sec:related}

\noindent
In traditional \cnn, convolutional and non-linearity computational units are alternated with subsampling operations. 
The purpose of subsampling is to increase the network receptive field while decreasing the feature maps sizes. 
A big receptive field is necessary for the network to get bigger context for the final prediction while the feature maps size reduction is a beneficial side effect allowing to increase the number of feature maps without overloading the (GPU) memory.
In the case of semantic segmentation where a full-resolution prediction is expected, the subsampling operators are detrimental as they decrease the final output resolution.

To get a prediction at the same resolution than the input image, Long, Shelhamer~\etal proposed recently Fully Convolutional Networks (FCN)~\cite{fully_convolutional} by adding a deconvolution part after a classical convolutional neural network. The idea is that, after decreasing in the convolutional network, a deconvolution part, using upsampling operator and deconvolution (or fractionally-strided convolution) increases the feature maps size back to the input resolution. 
Noh \etal \cite{learning_deconv} extended this idea by using maximum unpooling upsampling operators in the deconvolution part. The deconvolution network is the symmetric of the convolution one and each maximum pooling operation in the convolution is linked to a maximum unpooling one in the deconvolution by sharing the pooling positions. 
Ronneberger \etal \cite{u_net} are going even further with their U-Net by concatenating the feature maps obtained in the convolution part with feature maps of the deconvolution part to allow a better reconstruction of the segmented image. 
Finally, Lin \etal \cite{refine_net} used the same idea of U-Net but instead of concatenating the feature maps directly, they used a refineNet unit, containing residuals units, multi-resolutions fusions and chained residual pooling, allowing the network to learn a better semantic transformation.

All of these networks are based on the idea that subsampling is important to increase the receptive field and try to override the side effect of resolution loss with deconvolutionnal technics. 
In our \gridnet, composed of multiple streams working at different feature map sizes, we use the subsampling and upsampling operators as connectors between streams allowing the network to take decisions at any resolution.
The upsampling operators are not used to correct this side effect but to allow multi-scale decisions in the network.
In a recent work, Newell~\etal \cite{hourglass} stacked many U-Net showing that successive steps of subsampling and upsampling are important to improve the performance of the network. This idea is improved in our \gridnet with the strong connections between streams. 

Yu~\etal \cite{dilated_convolution} studied another approach to deal with the side effect of subsampling. 
They show that, for a semantic labelling task, the pooling operations are harmful. Therefore, they remove the subsampling operators to keep the feature maps at the same input resolution. Without subsampling, the receptive field is very small so they use dilated convolution to increase it. Contrarily to classical convolutions, where the convolution mask is applied onto neighbourhood pixels, dilated convolutions have a dilatation parameter to apply the mask to more and more apart pixels. 
In their work Wu~\etal \cite{wider_or_deeper} adapt the popular ResNet \cite{residual} pre-trained on ImageNet \cite{imagenet} for semantic segmentation.
ResNet \cite{residual} are very deep networks trained with residual connections allowing the gradient to propagate easily to the first layers of the network correcting the vanishing gradient problems.
Wu~\etal only keep the first layers of ResNet and change the classical convolutions into dilated ones. For memory problems, they also keep $3$ subsampling operators so the final output prediction is at $1/8$ of the input size, and then use linear interpolations to retrieve the input resolution. In \cite{pyramid_scene_parsing}, Zhao~\etal replace the linear interpolation by a Pyramid Pooling module. The pyramid pooling module is composed of multiple pooling units of different factors, followed by convolutions and upsample operators to retrieve the original size. All the feature maps obtained with different pooling sizes are then concatenated  before a final convolution operator that gives the prediction. When Zhao~\etal add a module at the end of the network to increase the feature maps size and allow a multi-scale decision, we incorporate this multi-scale property directly into our network with the different streams. 

In their work, He~\etal \cite{identity_mappings} study the importance of residual units and give detailed results on the different strategies to use residual connections (whether batch normalisation should be used before the convolutions, whether linearity operator should be used after the additions, etc.). \gridnet also benefits from these residuals units. 

With their Full Resolution Residual Network (FRRN) \cite{full_resolution_residual}, Pohlen~\etal combine a conv-deconv network with a residual one. They also use different streams but only two of them: one for the residual network linked with upsampling and subsampling operations, and one for the conv-deconv network which does not have any residual connections. \gridnet subsumes FRNN and can be seen as a generalisation of this network. 

The idea of networks with multiple paths is not new \cite{interlinked_cnn,multi_scale_dense_cnn,neural_fabric}.
Zhou~\etal studied a face parsing task with interlinked convolutional neural networks~\cite{interlinked_cnn}. 
An input image is used at different resolutions by multiple CNN whose feature maps are interconnected.
Huand~\etal~\cite{multi_scale_dense_cnn} use the same architecture but make it dynamically adaptable to computational resource limits at test time.
Recently, Saxena~\etal have presented Convolutional Neural Fabrics~\cite{neural_fabric} which structure forming a grid is very similar to ours and which also use the full multi-scale grid to make predictions. However, to better scale to full resolution images and large size datasets, we make use of residual units and we introduce a new dropout technique to better train our grids. Besides, we constrain our network, similarly to conv-deconv ones, to have down-sampling layers, followed by upsampling blocks, where ~\cite{neural_fabric} use up and down sampling across all network layers.

\section{\gridnet}
\label{sec:gridnet}

\begin{figure*}[t]
\centering
\includegraphics[width=0.99\textwidth]{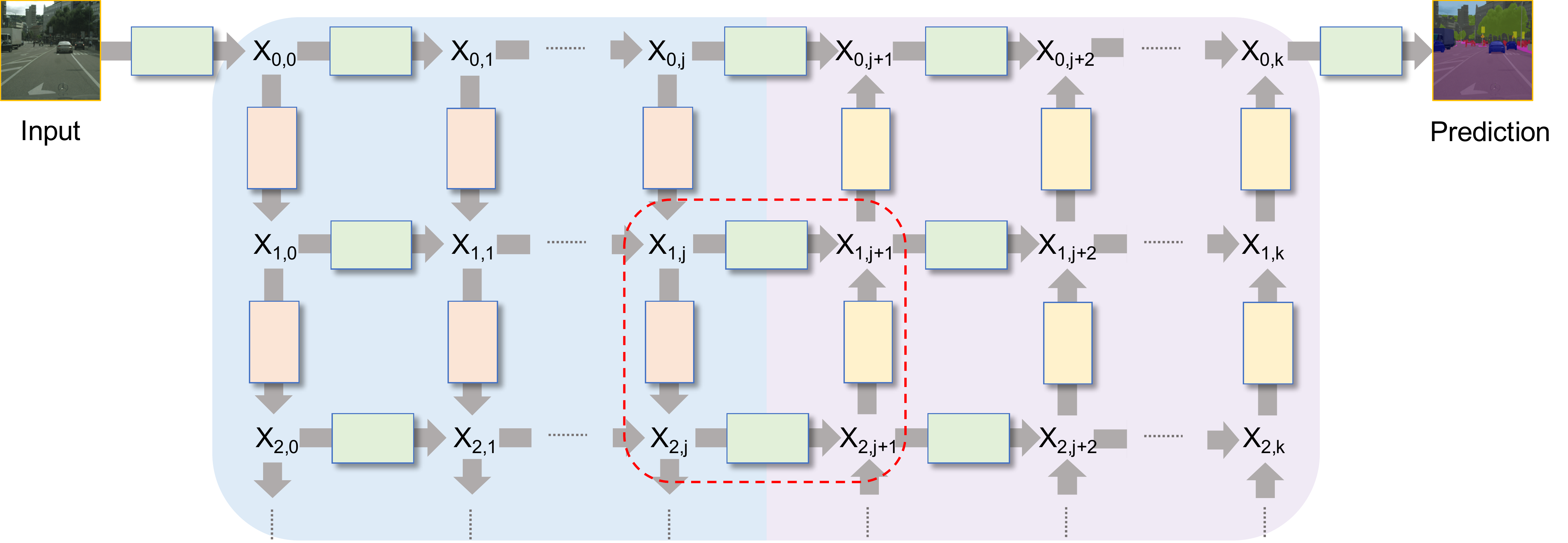}
\caption{\label{fig:gridnet} \gridnet: each green unit is a residual bloc, which does not change the input map resolution nor the number of feature maps. Red blocks are convolutional units with resolution loss (subsampling) and twice the number of feature maps. Yellow units are deconvolutional blocks which increase the resolution (upsampling) and divide by two the number of feature maps. A zoom on the red square part with a detailed compositions of each blocks is shown in Figure~\ref{fig:block}}
\end{figure*}


\begin{SCfigure}
\centering
\caption{\label{fig:block} Detailed schema of a GridBlock. Green units are residual units keeping feature map dimensions constant between inputs and outputs. Red units are convolutional +  subsampling and increase the feature dimensions. Yellow units are deconvolutional + upsampling and decrease the feature dimensions (back to the original one to allow the addition).
Trapeziums illustrate the upsampling/subsampling operations obtained with strided convolutions.
BN=Batch Normalization.}
\includegraphics[width=0.5\textwidth]{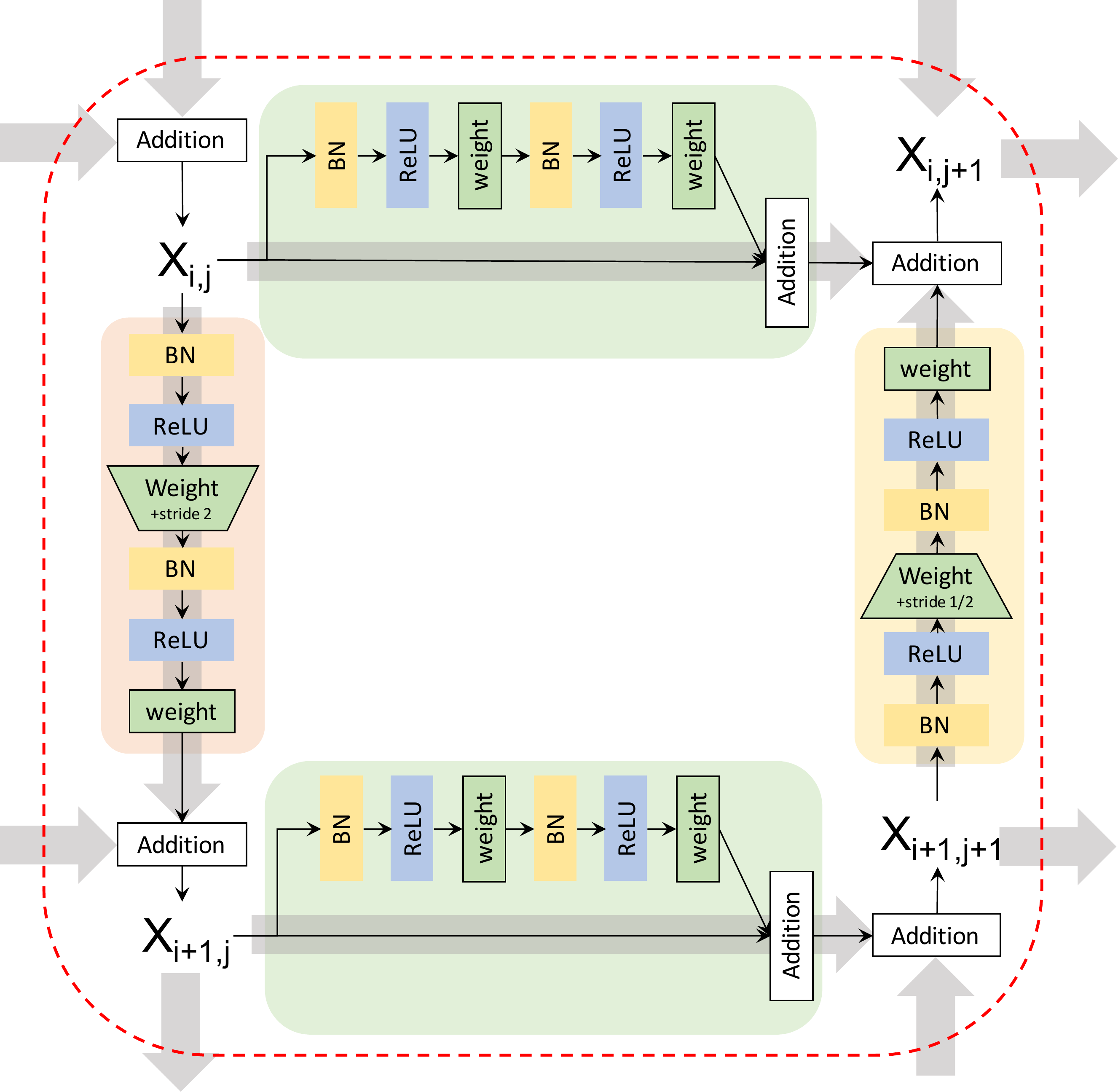}
\end{SCfigure}

\noindent
The computation graph of \gridnet is organised into a two-dimensional grid pattern, as shown in Figure \ref{fig:gridnet}. Each feature map $X_{i,j}$ in the grid is indexed by line $i$ and column $j$. Maps are connected through computation layers. Information enters the model as input to the first block of line 0 and leaves it as output from the last block of line 0. Between these two points, information can flow in several paths, either directly between these entry/exit points in a straight line or in longer paths which also involve lines with indexes $\neq 0$.

Information is processed in layers which connect blocks $X_{i,j}$. The main motivation of our model is the difference between layers connecting feature maps horizontally or vertically:
We call horizontal connections ``\emph{streams}''. Streams are fully convolutional and keep feature map sizes constant. They are also residual, i.e. they predict differences to their input \cite{residual}. Stream blocks are green in Figure~\ref{fig:gridnet}.
Vertical computing layers are also convolutional, but they change the size of the feature maps: according to the position in the grid, spatial sizes are reduced by subsampling or increased by upsampling, respectively shown as red and yellow blocks in Figure~\ref{fig:gridnet}. Vertical connections are \emph{NOT} residual.
The main idea behind this concept is an adaptive way to compute how information flows in the computation graph. Subsampling and upsampling are important operations in resolution preserving networks, which allow to increase the size of the receptive fields significantly without increasing filter sizes, which would require a higher number of parameters\footnote{An alternative would be to use dilated convolutions with the \emph{\`a trous} algorithm \cite{dilated_convolution}.}. On the other hand, the lost resolution needs to be generated again through learned upsampling layers. In our network, information can flow on several parallel paths, some of which preserve the original resolution (horizontal only paths) and some of which pass through down+up sampling operations. In the lines of the skip-connections in U-networks \cite{u_net}, we conjecture that the former are better suited for details, whereas high-level semantic information will require paths involving vertical connections.

Following the widespread practice, each subsampling unit reduces feature map size by a factor $2$ and multiplies the number of feature maps by $2$. More formally, if the stream $X_i$ takes as input a tensor of dimension $(F_i~\times~W_i~\times~H_i)$ where $F_i$ is the number of feature maps and $W_i$, $H_i$ are respectively the width and height of the map, then the stream $X_{i+1}$ is of dimension $(F_{i+1}~\times~W_{i+1}~\times~H_{i+1}) = (2F_i~\times~W_i / 2~\times~H_i / 2)$.


Apart from border blocks, each feature map $X_{i,j}$ in the grid is the result of two different computations: one horizontal residual computation processing data from $X_{i,j-1}$ and one vertical computation processing data from $X_{i-1,j}$ or $X_{i+1,j}$ depending if the column is a subsampling or upsampling one. Several choices can be taken here, including concatenating features, summing, or learning a fusion. We opted for summing, a choice which keeps model capacity low and blends well with the residual nature of the grid streams. The details are given as follows: let $\Theta^{Res}(.)$, $\Theta^{Sub}(.)$ and $\Theta^{Up}(.)$ be respectively the mapping operation for the residual unit (green block in Figure~\ref{fig:gridnet}), subsampling unit (red block) and upsampling unit (yellow block). Each mapping takes as input a feature tensor $X$ and some trainable parameters $\theta$. 

If the column $j$ is a subsampling column then:
$$X_{i,j} = \highlight[mygreen]{X_{i,j-1} + \Theta^{Res}(X_{i,j-1},\theta^{Res}_{i,j})} + \highlight[myred]{\Theta^{Sub}(X_{i-1,j},\theta^{Sub}_{i,j})}$$

Otherwise, if the column $j$ is an upsampling one then:
$$X_{i,j} = \highlight[mygreen]{X_{i,j-1} + \Theta^{Res}(X_{i,j-1},\theta^{Res}_{i,j})} + \highlight[myyellow]{\Theta^{Up}(X_{i+1,j},\theta^{Up}_{i,j})}$$
Border blocks are simplified in a natural way.
An alternative to summing is feature map concatenation, which increases the capacity and expressive power of the network. Our experiments on this version showed that it is much more difficult to train, especially since it is trained from scratch.

The capacity of a \gridnet is defined by three hyper parameters, $N_S$, $N_{Cs}$ and $N_{Cu}$ respectively the number of residual streams, the number of subsampling columns and the number of upsampling columns. Inspired by the symmetric conv-deconv networks \cite{fully_convolutional}, we set $N_{Cs}{=}N_{Cu}$ in our experiments, but this constraint can be lifted.

\begin{figure*}[h]
\centering
\includegraphics[width=0.7\textwidth]{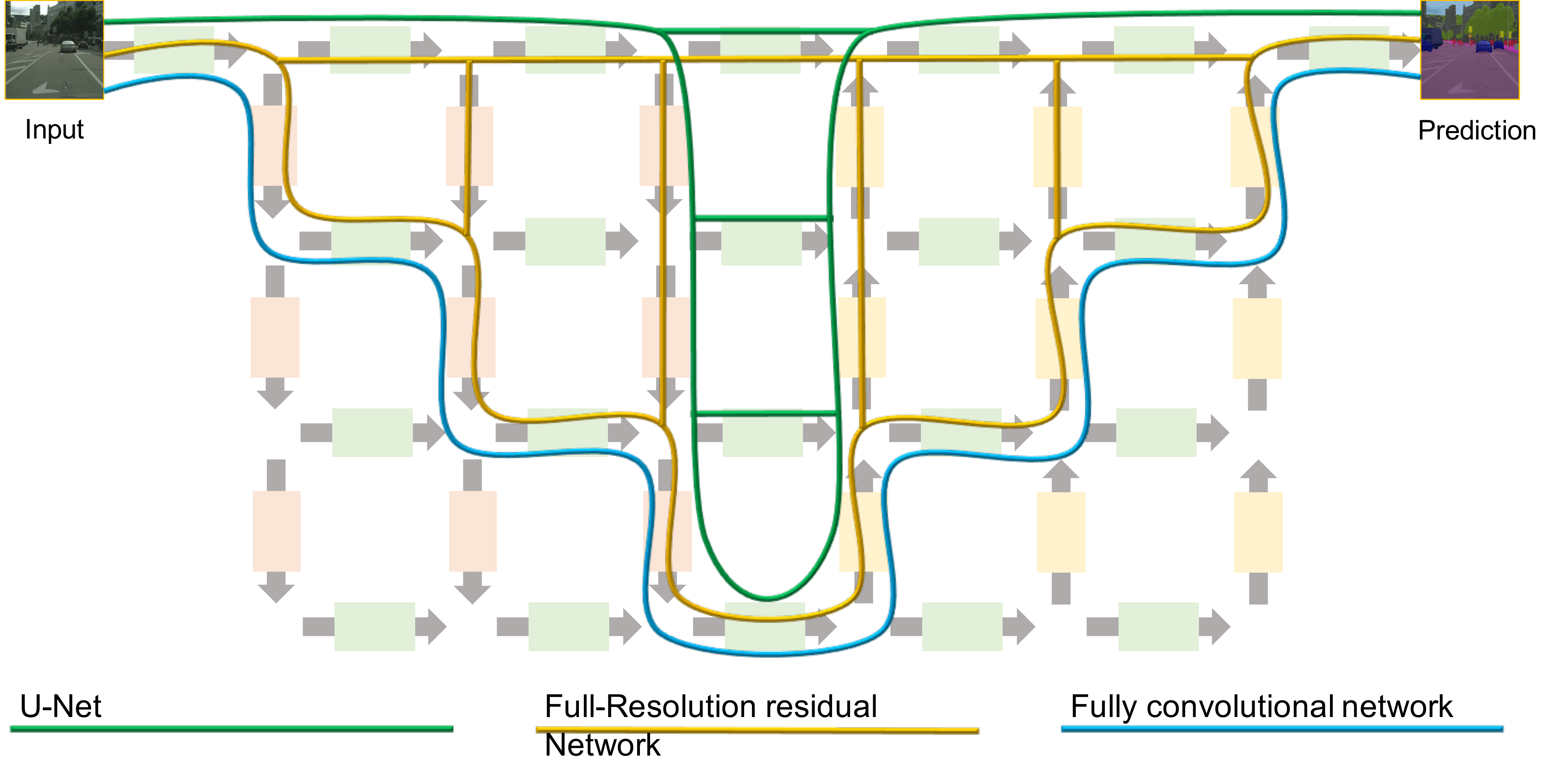}
\caption{\label{fig:generalisation} \gridnets generalize several classical resolution preserving neural models such as conv-deconv networks \cite{fully_convolutional} (blue connections), U-networks \cite{u_net} (green connections) and Full Resolution Residual Networks (FRRN) \cite{full_resolution_residual} (yellow connections).}
\end{figure*}

\gridnet generalize several classical resolution preserving neural models, as shown in Figure \ref{fig:generalisation}. Standard models can be obtained by removing connections between feature maps in the grid. If we keep the connections shown in blue in Figure \ref{fig:generalisation}, we obtain conv-deconv networks \cite{fully_convolutional} (a single direct path). U-networks \cite{u_net} (shown by green connections) add skip-connections between down-sampling and corresponding up-sampling parts, and Full Resolution Residual Networks (FRRN) \cite{full_resolution_residual} (shown as yellow connections) add a more complex structure.

\subsection{Blockwise dropout for \gridnets}  
\label{sec:dropout}

\begin{figure*}[h]
\centering
\includegraphics[width=0.7\textwidth]{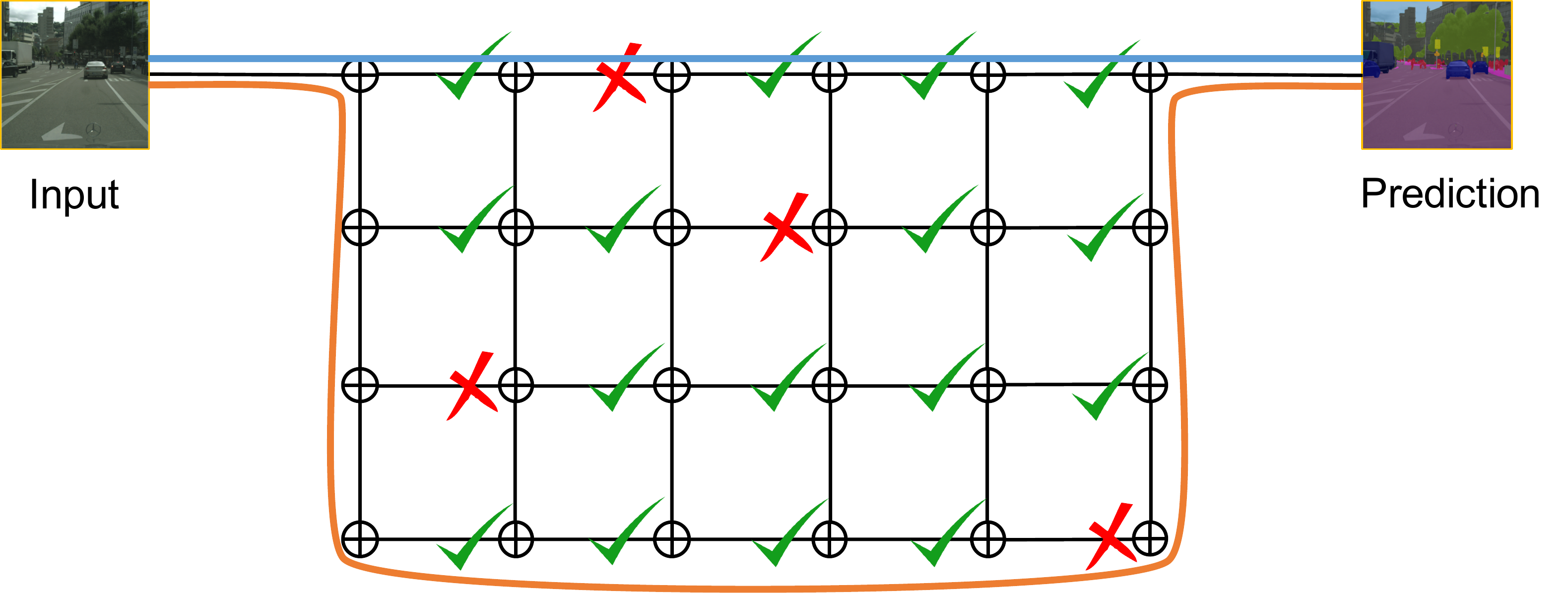}
\caption{\label{fig:dropout} The blue path only using the high resolution stream is shorter than the orange path which also uses low resolution streams. To force the network to use all streams we randomly drop streams during training, indicated by red crosses.}
\end{figure*}

A side effect of our 2D grid topology with input and output both situated on line 0 is that the path from the input to the output is shorter across the high resolution stream (blue path in figure~\ref{fig:dropout}) than with the low resolution ones (e.g. the orange path in Figure~\ref{fig:dropout}). Longer paths in deep networks may fall into the well known problems of vanishing gradients. As a consequence, paths involving lower resolution streams take more time to converge and are generally more difficult to train.
To force the network to use all of its available streams, we employed a technique inspired by dropout, which we call  \emph{total dropout}. It consists in randomly dropping residual streams and setting the corresponding residual mappings to zero.

More formally, let $r_{i,j} = Bernoulli(p)$ be a random variable taken from a \emph{Bernoulli} distribution, which is equal to $1$ with a probability $p$ and $0$ otherwise. Then, the feature map computation becomes:
$X_{i,j} = X_{i,j-1} + r_{i,j}(\Theta^{Res}(X_{i,j-1},\theta^{Res}_{i,j}))  + \Theta^{\{Sub;Up\}}(X_{i\pm1,j},\theta^{\{Sub;Up\}}_{i,j})$

\subsection{Parameter count and memory footprint for \gridnets}

In neural networks, the memory footprint depends on both the number of activations and the number of trainable parameters.
In many architectures, these two numbers are highly correlated.
While it is still the case in a \gridnet, the grid structure provides a finer control over these numbers.
Let us consider a \gridnet built following the principles from Section~\ref{sec:gridnet}:
with $N_S$ streams, $N_{Cs}$ subsampling columns and $N_{Cu}$ upsampling columns, with the first stream having $F_0$ feature maps at resolution $W_0{\times}H_0$, and the others streams obtained by downsampling by $2{\times}2$ and increasing the feature maps by 2.
From the exact computation of the number of parameters $nb_{param}$ and the number of activation values $nb_{act}$, we can derive meaningful approximations:
$$
nb_{param} \approx 18 \times 2^{2*(N_s-1)} \, F_0^2 \,  (2.5 N_{Cs} + N_{Cu} - 2)
$$

This approximation illustrates that the number of parameters is most impacted by the number of streams $N_S$, followed by the number of feature maps (controlled by $F_0$), and only then, by the number of columns. 
$$
nb_{activ} \approx 6 \, H_0 \, W_0 \, F_0 \, \left( 4 N_{Cu} + 3 N_{Cs} - 2 \right)
$$

This shows that the number of activations mainly depends on the first stream size (width, height and number of feature maps) and grows linearly with the number of columns.
In practice, the total memory footprint of a network at training time depends not only on its number of parameters and on the number of activations, but also on both the choice of the optimizer and on the mini-batch size.
The gradient computed by the optimizer requires the same memory space as the parameters themselves and the optimizer may also keep statistics on the parameters and the gradients (as does Adam).
The mini-batch size mechanically increases the memory footprint as the activations of multiple inputs need to be computed and stored in parallel.

\section{Experimental results}
\label{sec:exp}

We evaluated the method on the Cityscapes dataset \cite{Cordts2016Cityscapes}, which consists in high resolution ($1024\times2048$ pixels) images taken from a car driving across $50$ different cities in Germany. 
$2975$ training images and $500$ test images have been fully labelled with $30$ semantic classes. However, only $19$ classes are taken into account for the automatic evaluation on the Cityscapes website \footnote{https://www.cityscapes-dataset.com/}, therefore we trained \gridnet on these classes only. Semantic classes are also grouped into $8$ semantic categories.
The ground truth is not provided for the test set but an online evaluation is available on the Cityscapes website. The dataset contains also $19998$ images with coarse (polygonal) annotations but, we chose not to use them for training because they increase the unbalance ratio of the label distribution which is harmful to our performance measures.

The Cityscapes performance are evaluated based on the Jaccard Index, commonly known as the Pascal VOC Intersection-over-Union (IoU) metric. 
The IoU is given by $\frac{TP}{TP+FP+FN}$ where $TP$, $FP$ and $FN$ are the number of True Positive, False Positive and False Negative classified pixels. 
IoU is biased toward object instances that cover a large image area so, an instance-level intersection-over-union metric \emph{iIoU} is also used. The \emph{iIoU} is computed by weighting the contribution of each pixel by the ratio of the class average instance size, to the size of the respective ground truth instance. 
Finally, they give results accuracy for two semantic granularities (class and category) with the weighted and not weighted IoU metric leading to $4$ measurements.

We tested \gridnet with $5$ streams with the following feature map dimensions 16, 32, 64, 128 and 256. \gridnet is composed of $3$ subsampling columns (convolutional parts) followed by $3$ upsampling columns (deconvolutional parts).
This "$5$ streams / $6$ columns" configuration provides a good tradeoff between memory consumption and number of parameters: the network is deep enough to have a good modelling capacity with few enough parameters to avoid overfitting phenomena. This configuration allows us to directly fit in our GPU memory a batch of 4 $400\times400$ input images. As a consequence, the lowest resolution stream deals with feature maps of size ($256~\times~25~\times~25$).

We crop patches of random sizes (between $400\times400$ and $1024\times1024$) at random locations in the high resolution input images ($1024\times2048$).  All the patches are resized to $400\times400$ and fed to the network. 
For data augmentation, we also apply random horizontal flipping. 
We do not apply any post-processing for the images but we added a batch normalization layer at the input of the grid. 
We use the classical cross-entropy loss function to train our network using the Adam optimizer with a learning rate of $0.01$, a learning rate decay of $5\times10^{-6}$, $\beta_1 = 0.9$, $\beta_2 = 0.999$ and an $\epsilon = 1\times10^{-8}$.
After $800$ epochs, the learning rate is decreased to $0.001$. We stopped our experiments after $10$ days leading to approximately $1900$ training epochs. 
For testing we fed the network with images at resolutions $\frac{1}{1},\frac{1}{1.5},\frac{1}{2},\frac{1}{2.5}$ and used a majority vote over the difference scale for the final prediction.  

\subsection{Discussion}
\label{se:results}

We conducted a study to evaluate the effects of each of our architectural components and design choices. The results are presented in Table~\ref{res:exp} and ~\ref{res:structure}. 

In Table~\ref{res:exp}, Sum$\dagger$ is the results given by the network presented in section~\ref{sec:gridnet} with total dropout operators (see section~\ref{sec:dropout}). Total dropout proved to be a key design choice, which lead to significative improvement in accuracy.
We also provide results of a fully residual version of GridNet, where identity connections are added in both horizontal and vertical computing connections (whereas the proposed method is residual in horizontal streams only). Full residuality did not prove to be an advantage. Total dropout did not solve learning difficulties and further impacted training stability negatively.
Finally, concatenation of horizontal and vertical streams, instead of summing, did also not prove to be an optimal choice. We conjecture that the high capacity of the network did not prove to be an advantage.


\begin{table}[t]
\begin{center}
\scriptsize
\begin{tabular}{lccc || c c | c c }
&&&& \multicolumn{4}{c}{Performance measures} \\
Fusion & h-residual & v-residual & Total dropout & IoU class  & iIoU class  & IoU categ. & iIoU categ.  \\
\hline
Sum    &  &   & \bcmark  & 60.2 &34.5 &83.9 & 67.3 \\
Sum    & \bcmark &   &   & 57.2 & 35.6 & 83.1 & 68.4 \\
Sum$\dagger$    & \bcmark &   & \bcmark & {\bf 65.0} & {\bf 43.2} & 85.6 & 70.1 \\
Sum    & \bcmark & \bcmark &   & 57.6 & 36.8 & {\bf 86.0} & {\bf 72.6} \\
Sum    & \bcmark & \bcmark & \bcmark & 35.6 & 23.0 & 62.1 & 60.3 \\
Concat & \bcmark &   &   & 53.9 & 34.0 & 82.2 & 65.2\\
\end{tabular}
\end{center}
\caption{\label{res:exp} Results of different \gridnet variants on the Cityscapes validation set: "Fusion" indicates how feature maps are fused between horizontal and vertical blocks. The second and third columns indicate whether horizontal (resp. vertical) computations are residual. $\dagger$ stands for the final proposed method.}
\end{table}

\begin{table}[t]
\begin{center}
\scriptsize
\begin{tabular}{ c c || c c | c c }
& & \multicolumn{4}{c}{Performance measures} \\
Nb columns & Nb Features maps per streams  & IoU class  & iIoU class  & IoU categ. & iIoU categ.  \\
\hline
8 & \{8, 16, 32, 64, 128\} & 57.5	 & 40.0 & 83.8 & {\bf 71.8} \\
16 &  \{8, 16, 32, 64, 128\} & 56.3& 38.6 & 82.3 & 70.2 \\
8 & \{8, 16, 32, 64, 128, 256, 512\} & 59.2 & 41.1 & 83.5 & 71.0 \\
12 & \{8, 16, 32, 64, 128, 256, 512\} & {\bf 59.5} & {\bf 41.7} & {\bf 84.0} & 70.9 \\
\end{tabular}
\end{center}
\caption{\label{res:structure} Results of the impact of different number of columns and streams. No data augmentation (only one scale) was use in testing.}
\end{table} 

Table~\ref{res:structure} presents the impact of the number of columns and streams used in \gridnet.
We started with a \gridnet composed of 8 columns (4 subsampling followed by 4 upsampling) and 5 streams (results using networks with other configurations of the subsampling/upsampling units are presented in Table \ref{tab:network}).
Instead of using 16 feature maps in the first stream, we used only 8 to reduce the memory consumption and allow us to increase the number of columns and/or streams while still coping with our hardware constraints.
Networks are trained until convergence and the tests are performed without data augmentation (only one scale and no majority vote). From Table~\ref{res:structure}, we can see that increasing the number of streams increases the performance (from 57.5 to 59.2 for the IoU class accuracy), but increasing only the number of columns (from 8 to 16) do not improve the accuracy while increasing the training complexity. A low number of streams limits the abstraction power of the network. Increasing both the number of streams and of the columns (up to the hardware capacity), improves all the performance measures.

\subsection{Qualitative and Quantitative Results}
\label{sec:results}

\begin{figure}[t]
\centering
\includegraphics[width=0.845\textwidth]{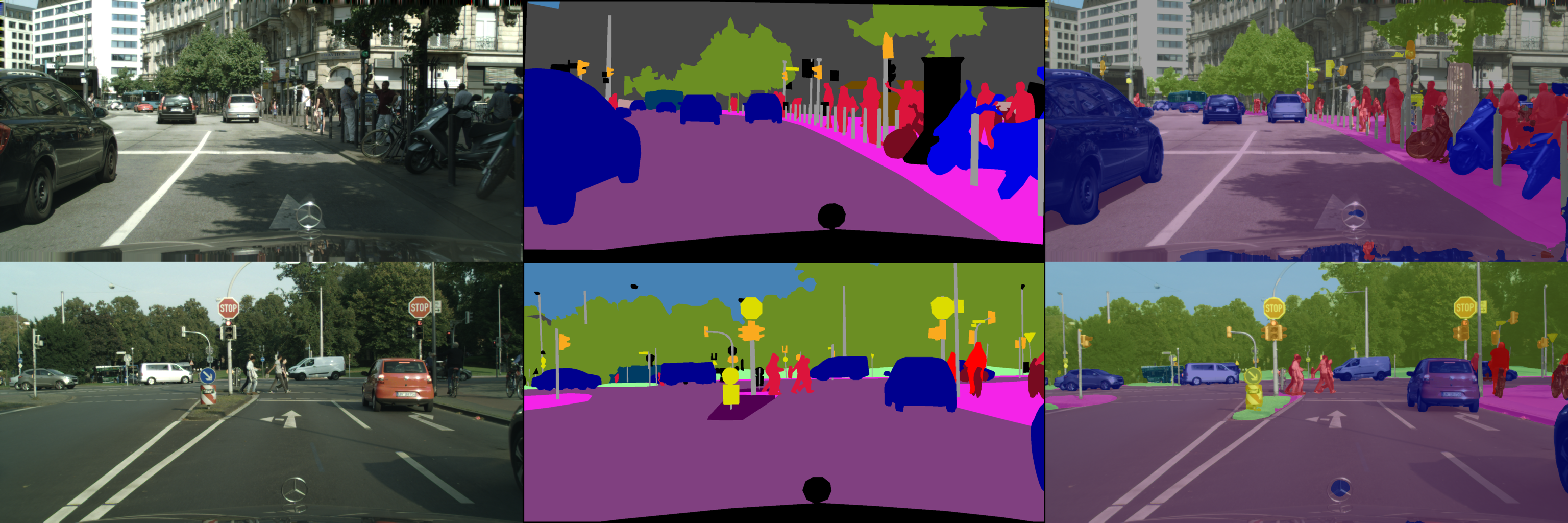}
\caption{\label{img:results} Semantic segmentation results obtained with \gridnet. On the left, he input image, in the middle the ground truth and on the right, our results.}
\end{figure}

Figure \ref{img:results} shows segmentation results of some sample images. In Table \ref{tab:network}, we compare the results of our \gridnet compared to state-of-the-art results taken from the official Cityscapes website. We restrict the comparison to methods that the same input information as us (no coarse annotations, no stereo inputs). Our network gives results comparable with the state-of-the-art networks, in particular, the FRNN network presented in Section \ref{sec:related}. 

All other results on the Cityscapes website have been obtained by networks pre-trained for classification using the Imagenet dataset. Nevertheless, among the 9 other reported results, only one of them (RefineNet) give slightly better results than our network. 

\begin{table}[h]
\begin{center}
\scriptsize
\begin{tabular}{c | c || c c | c c}
 \multirow{2}{*}{Name} & Trained  & \multicolumn{4}{c}{Performance measures} \\ 
 &  from scratch & IoU class  & iIoU class  & IoU categ. & iIoU categ. \\
\hline
FRRN - \cite{full_resolution_residual}                                          & \cmark   & {\bf 71.8}  & {\bf 45.5}  & {\bf 88.9}  & {\bf 75.1}\\
\rowcolor{Gray}
GridNet                                      & \cmark   & 69.45 & 44.06 & 87.85 & 71.11\\
\rowcolor{Gray}
GridNet - Alternative                                     & \cmark   & 66.8 & 38.24 & 86.55 & 68.98 \\
\hline
RefineNet - \cite{refine_net}                                      & \xmark    & {\bf 73.6}  & 47.2  & 87.9  & 70.6\\
Lin~\etal - \cite{piecewise}                                & \xmark    & 71.6  & {\bf 51.7}  & 87.3  & 74.1\\
LRR - \cite{lrr}                                          & \xmark    & 69.7  & 48    & {\bf 88.2}  & {\bf 74.7}\\
Yu~\etal - \cite{dilated_convolution}                                   & \xmark    & 67.1  & 42    & 86.5  & 71.1\\
DPN - \cite{dpn}                                             & \xmark    & 66.8  & 39.1  & 86    & 69.1\\
FCN - \cite{fully_convolutional}                                       & \xmark    & 65.3  & 41.7  & 85.7  & 70.1\\
Chen~\etal - \cite{DeepLab}                         & \xmark    & 63.1  & 34.5  & 81.2  & 58.7\\
Szegedy~\etal - \cite{deeper_convolution}                                  & \xmark    & 63    & 38.6  & 85.8  & 69.8\\
Zheng~\etal - \cite{crfasrnn}                                        & \xmark    & 62.5  & 34.4  & 82.7  & 66  \\
\end{tabular}
\end{center}
\caption{\label{tab:network} Results on the Cityscapes dataset benchmark. We only report published papers which use the same data as us (no coarse annotations, no stereo inputs). "GridNet - Alternative" is another structure closer to~\cite{neural_fabric} where up and down sampling columns are interleaved.}
\end{table}

\section{Conclusion}
\label{sec:conc}

\noindent
We have introduced a novel network architecture specifically designed for semantic segmentation. The model generalizes a wide range of existing neural models, like conv-deconv networks, U-networks and Full Resolution Residual Networks. A two-dimensional grid structure allows information to flow horizontally in a residual resolution-preserving way or vertically through down- and up-sampling layers. \gridnet shows promising results even when trained from scratch (without any pre-training). We believe that our network could also benefit from better weight initialisation, for example by pre-training it on the ADE20K dataset. 

\section*{Acknowledgment}
Authors acknowledge the support from the ANR project SoLStiCe (ANR-13-BS02-0002-01).
They also want to thank Nvidia for providing two Titan X GPU.

\end{document}